\pdfoutput=1

\documentclass[11pt]{article}

\usepackage{acl}

\usepackage{times}
\usepackage{latexsym}

\usepackage[T1]{fontenc}

\usepackage[utf8]{inputenc}

\usepackage{microtype}

\usepackage{amsfonts}

\usepackage{inconsolata}

\usepackage{graphicx}
\usepackage{hyperref}
\usepackage{url}
\usepackage{algorithm}
\usepackage{algorithmic}
\usepackage{amsmath}

\usepackage{colortbl}
\usepackage{multirow}
\usepackage{multicol}
\usepackage{float}
\usepackage{booktabs}
\usepackage{tcolorbox}

\title{SSH: Sparse Spectrum Adaptation via Discrete Hartley Transformation}

\author{
 \textbf{Yixian Shen}, 
    \textbf{Qi Bi}, 
    \textbf{Jia-Hong Huang}, 
    \textbf{Hongyi Zhu}, 
    \textbf{Andy D. Pimentel}, 
    \textbf{Anuj Pathania} \\
    University of Amsterdam, Amsterdam, the Netherlands \\
\\
 \texttt{\{y.shen, q.bi, j.huang, h.zhu, a.d.pimentel, a.pathania\}@uva.nl}
}

\begin{document}
\maketitle
\begin{abstract}
Low-rank adaptation (LoRA) has been demonstrated effective in reducing the trainable parameter number when fine-tuning a large foundation model (LLM). However, it still encounters computational and memory challenges when scaling to larger models or addressing more complex task adaptation.
In this work, we introduce \textbf{S}parse \textbf{S}pectrum Adaptation via Discrete \textbf{H}artley Transformation (SSH), a novel approach that significantly reduces the number of trainable parameters while enhancing model performance. 
It selects the most informative spectral components across all layers, under the guidance of the initial weights after a discrete Hartley transformation (DHT). 
The lightweight inverse DHT then projects the spectrum back into the spatial domain for updates. 
Extensive experiments across both single-modality tasks—such as language understanding and generation—and multi-modality tasks—such as visual-text understanding—demonstrate that SSH outperforms existing parameter-efficient fine-tuning (PEFT) methods while achieving 
substantial reductions in computational cost and memory requirements. 
\end{abstract}

\section{Introduction}

\begin{figure}[!t]
    \centering
    \includegraphics[width=1\linewidth]{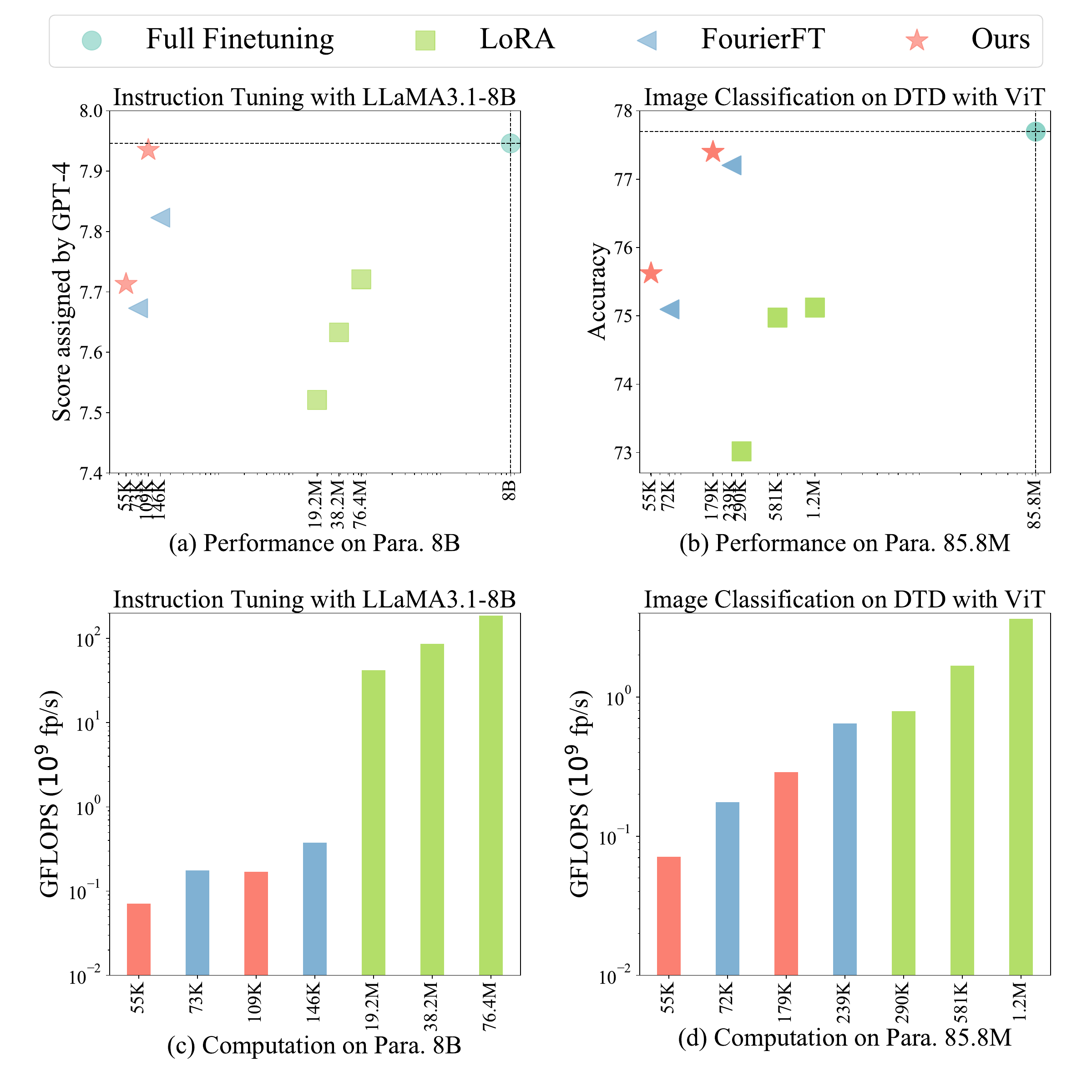}
    \caption{ \small Performance and computation comparison of fine-tuning methods in NLP and CV Tasks. (a) For NLP on LLaMA3.1-8B, SSH achieves 7.93 GPT-4 score, closely matching full fine-tuning's 7.95 score, while using less than 0.1\% of the parameters. (b) In CV tasks, SSH achieves 77.4\% accuracy, matching the performance of full fine-tuning with significantly fewer parameters. (c) \& (d) SSH reduces up to 55\% of GFLOPs compared to FourierFT in both NLP and CV tasks, providing significant computation efficiency gains.
    }
    \label{fig:mov}
\end{figure}

Pretrained foundation models, such as GPT-4~\cite{islam2024gpt}, LLaMA3.1~\cite{touvron2023llama}, and Vision Transformers (ViT)\cite{dosovitskiy2020image}, have demonstrated remarkable performance across diverse natural language processing (NLP)\cite{huang2024gradient,tao2024robustness,shen2024comparative,shen2024altgen,shen2024parameter} and vision tasks~\cite{huang2025image2text2image,huang2024novel}. 
This success can largely be attributed to the unprecedented growth in model size~\cite{wei2022emergent}. 
However, as these models scale up to billions of parameters, adapting them for downstream tasks in various domains presents significant computational~\cite{shen2023thermal,shen2023thermal1,shen2022tcps,niknam20233d} and memory challenges~\cite{guo2024easter,aghapour2024piqi}. 
Fully fine-tuning these large models becomes prohibitively expensive, both in terms of memory consumption and computational resources. 
For instance, fine-tuning the LLaMA3.1 model with 8 billion parameters requires 60GB of GPU memory. 
While parameter-efficient fine-tuning (PEFT) methods like LoRA~\cite{hu2022lora} and QLoRA~\cite{dettmers2024qlora} can reduce memory requirements to 16GB and 6GB, respectively, they still fall short when applied to larger models such as LLaMA3.1 70B, which demands up to 48GB of memory, even with Q-LoRA.

LoRA and its successors~\cite{lialin2023relora,renduchintala2024tiedlora,liu2024dora} have made substantial progress by introducing low-rank adaptations and quantization techniques to mitigate the memory overhead when fine-tuning large models. 
Despite their ability to reduce the number of trainable parameters, as these methods primarily operate in the weight space of the original models, they still face limitations in terms of overall computational efficiency and GPU memory requirements, especially when scaling to massive models.

A promising alternative to address this bottleneck is to leverage the frequency transformations, which offer a more compact representation of model weights with less trainable parameters. 
Recent work on frequency-based PEFT, such as the discrete Fourier transform (DFT) approach~\cite{gao2024parameter}, has shown that transforming weight matrices into the spectral domain and updating spectral components can significantly reduce the number of trainable parameters. However, DFT operates in the complex domain, introducing potential computational overhead and numerical instability~\cite{press2007numerical}, particularly in large-scale models~\cite{gao2024parameter}. 
Such numerical inaccuracies inevitably degrade the performance. Moreover, the asymmetry between DFT and its inverse (iDFT) complicates forward and backward transformations, increasing computational intensity. 

To address these challenges, we propose a novel fine-tuning framework, \textbf{S}parse \textbf{S}pectrum Adaptation via Discrete \textbf{H}artley Transform (SSH).
The advantage of leveraging DHT over DFT~\cite{gao2024parameter} is twofold.
Firstly, DHT integrates both cosine and sine components in a single operation. It avoids the imaginary number computation, simplifies the computations and improves the numerical stability.
Secondly, its symmetry—where the inverse transformation is identical to the forward—streamlines the fine-tuning process, allowing efficient transitions between time and frequency domains. 
SSH selectively fine-tunes the most critical frequency components, identifies through the energy compaction properties, and efficiently recovers weight updates via inverse DHT. By capitalizing on the symmetrical nature of DHT and reducing parameter usage, SSH provides a highly parameter-efficient and computationally optimized fine-tuning strategy.

As shown in Fig.~\ref{fig:mov}(a) \& (b), SSH achieves comparable performance in both instruction tuning with the LLaMA3.1-8B model and image classification with the ViT 85.8M model, with significantly fewer parameters than LoRA and FourierFT. 
Fig.~\ref{fig:mov}(c) \& (d) further demonstrate its superior computational efficiency, requiring 55\% fewer GFLOPs than FourierFT. 
This is due not only to reduce the parameter count but also to avoid the complex number handling required by FourierFT, where real and imaginary parts must be processed separately.

Concretely, our contribution is threefold.
\begin{itemize}
\item We introduce \textbf{SSH}, a novel PEFT method based on discrete Hartley transform.
It simplifies operations and enhances numerical stability by avoiding complex arithmetic.
\item An energy-based frequency selection strategy is proposed to help SSH selectively fine-tune the most critical frequency components.
\item SSH shows superior performance and computational efficiency across various NLP, vision, and multi-modal tasks, achieving significant parameter savings and GFLOPs reduction.
\end{itemize}
\section{Related Work}

\noindent \textbf{Low-Rank Adaptation} (LoRA) \cite{hu2022lora} reduces trainable parameters by learning low-rank matrices that bypass full-weight updates, minimizing memory usage for gradients and optimizers. 
Different from adapter-based methods \cite{he2021towards, pfeiffer2020adapterfusion, lin2020exploring, liao2023make, liao2023parameter}, LoRA incurs no inference overhead as the low-rank updates are merged with the frozen weights. However, scaling LoRA to larger models and more complex tasks remains challenging.
Recent improvements, including AdaLoRA \cite{zhang2303adaptive}, VeRA \cite{kopiczko2023vera}, QLoRA \cite{dettmers2024qlora} and DoRA \cite{liu2024dora}, optimize parameter allocation and weight decomposition but still face scalability challenges on larger models. 

\noindent \textbf{Frequency-based Spectrum Learning} has been used to reduce trainable parameters while preserving model capacity. Prior works~\cite{xu2020learning,tang2022rethinking,yang2016exact} showed the effectiveness of compact and sparse spectral representation learning. Gao et al.~\cite{gao2024parameter} applied the Fourier Transform to fine-tune a subset of spectral coefficients, highlighting the potential of sparse spectrum adaptation in large foundation models. However, the DFT introduces complex operations, and the asymmetry between the DFT and its inverse increases computational overhead.
SSH addresses these issues with the real-valued DHT, which eliminates complex arithmetic, reduces computational complexity, and enhances numerical stability through symmetric transforms. Additionally, SSH’s energy-based sparse selection further decreases trainable parameters, improving efficiency and scalability.

\noindent \textbf{DHT} has shown potential in deep learning for model compression and computational efficiency. For example, \cite{rani2024content} employed DHT in medical image retrieval, \cite{ma2021high} used it in single-pixel imaging for efficient data acquisition, and \cite{coutinho2021low} leveraged it for media image compression and recovery. These works highlight DHT’s ability to reduce parameters while maintaining performance. 

\begin{algorithm}[t!]
\footnotesize
\caption{\textbf{SSH} Algorithm}
\label{alg:SSH}
\textbf{Input}: Input tensor $x$, number of parameters $n$, scaling factor $\alpha$, input dimension $d_1$, output dimension $d_2$, energy ratio $\delta$, pretrained layer weights $W$\\
\textbf{Output}: Transformed tensor $h$
\begin{algorithmic}[1]
\STATE \textbf{Initialization:}
\STATE $W_{F} = \text{DHT}(base\_layer.weight)$  \textcolor{blue}{//DHT for weights, Eq(\ref{eq2})}
\STATE  \textcolor{blue}{// Select top-($n \times \delta $ ) frequencies by energy}
\STATE $n_{\text{select}} = n \times \delta $ $\leftarrow$ RankTopEnergyFreq($\mathcal{M}$)
\STATE \textcolor{blue}{// Randomly select the rest of frequencies}
\STATE $n_{\text{random}} = n \times (1-\delta) $ $\leftarrow$ RandomSelectFreq($\mathcal{M}$)

\STATE \textcolor{blue}{// Initialize spectral coefficients}
\STATE $\Delta \mathbf{H}$ $\leftarrow$ KaimingInitial()

\STATE \textbf{Forward Pass:}
\STATE \textcolor{blue}{// Set $n$ selected freq. trainable and froze ($d_1 \times d_2 - n$).}
\STATE Set $n$ frequencies $\leftarrow$ requireGrad(True)
\STATE \textcolor{blue}{// Compute $\Delta W_T$ using inverse DHT based on Eq(\ref{eq3})}
\STATE $\Delta W_T \leftarrow \text{DHT}^{-1}(\Delta \mathbf{H}) \times \alpha$
\STATE \textcolor{blue}{// Merge $\Delta W$ with base layer output} 
\STATE $h \leftarrow W + \Delta W_T$
\STATE \textbf{return} $h$
\end{algorithmic}
\end{algorithm}

\begin{figure*}
    \centering
    \includegraphics[width=\linewidth]{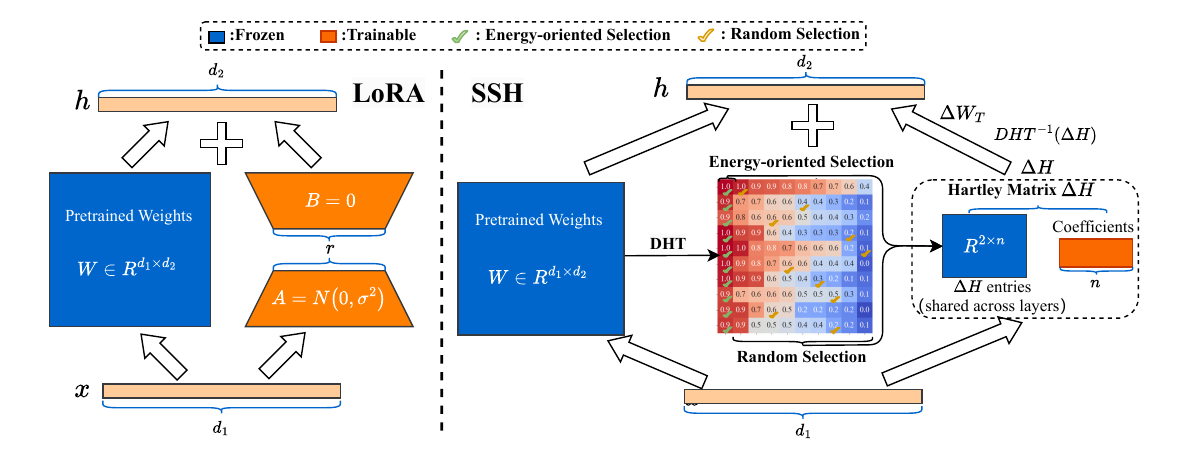}
    \caption{\small 
    Overview of \textbf{S}parse \textbf{S}pectrum Adaptation via Discrete \textbf{H}artley Transform (SSH). First, the discrete Hartley transform (DHT) is applied to the pretrained weights to extract and retain the most important frequency components. Then, a selective process identifies specific spectral coefficients to be learned as trainable parameters, which are organized into a spectral matrix. Finally, the modified spectral matrix is transformed back to the spatial domain through the symmetric application of the inverse discrete Hartley transform (iDHT), ensuring accurate reconstruction and efficient updates to the model's weights.
    }
    \label{fig:SSH}
\end{figure*}

\section{Methodology}

We propose SSH, a novel frequency selection strategy based on the discrete Hartley transform for PEFT, as illustrated in Fig.~\ref{fig:SSH}.  
It operates in the Hartley spectral domain and learns a set of spectral coefficients. 
Specifically, we employ the discrete Hartley transform to the pretrained weights, using \textit{energy-based frequency selection} to identify the most informative frequencies. In addition, we incorporate random sampling to ensure diversity in the selected frequencies while maintaining the representation ability.

The SSH algorithm~\ref{alg:SSH} selectively fine-tunes a pretrained layer's weights using the Discrete Hartley Transform (DHT) to capture the most significant frequency components. First, the layer's weights are transformed into the frequency domain using DHT, and the top frequencies, based on energy, are selected for fine-tuning, while the remaining frequencies are randomly chosen and kept frozen. The trainable spectral coefficients are initialized using Kaiming initialization, and during backpropagation, only the selected frequencies have their gradients updated. The inverse DHT is then applied to these updated spectral coefficients, scaled by a factor $\alpha$, to obtain the transformed weights in the spatial domain. These updates are merged with the original pretrained weights, resulting in the final transformed tensor. This approach ensures that only the most informative frequency components are fine-tuned, significantly reducing the number of trainable parameters while maintaining model performance.

\subsection{Sparse Hartley Spectral Learning}

Let $\mathbf{W}_0 \in \mathbb{R}^{d_1 \times d_2}$ denote the pretrained weight matrix, and $\Delta \mathbf{W} \in \mathbb{R}^{d_1 \times d_2}$ denote the weight change during fine-tuning. LoRA~\cite{hu2022lora,liu2024dora,gao2024parameter} models the weight change by low-rank decomposition, represented as $BA$, where $B \in \mathbb{R}^{d_1 \times r}$ and $A \in \mathbb{R}^{r \times d_2}$. The fine-tuned weight matrix $\mathbf{W}$ is expressed as:

\begin{small}
\begin{equation}
    \mathbf{W} = \mathbf{W}_0 + BA.
\end{equation}    
\end{small}

Rather than update the weights directly in the spatial domain, we project pretrained weights into the spectral domain using the 2D discrete Hartley transform, which helps us to select the most informative frequencies. 
DHT is an orthogonal transform similar to the discrete Fourier transform but operates solely on real-valued inputs, making it computationally efficient and suitable for fine-tuning. Given a weight matrix $\mathbf{W}_0 \in \mathbb{R}^{d_1 \times d_2}$, 
its frequency counterpart $\mathbf{H}_0$ after the 2D DHT is defined as:

\begin{small}
\begin{equation}
\label{eq2}
\begin{split}
    \mathbf{H}_0(u, v) &= \sum_{x=0}^{d_1-1} \sum_{y=0}^{d_2-1} \mathbf{W}_0(x, y) \times \\
    &\left[ \cos\left( \frac{2\pi ux}{d_1} + \frac{2\pi vy}{d_2} \right) - \sin\left( \frac{2\pi ux}{d_1} + \frac{2\pi vy}{d_2} \right) \right],
\end{split}
\end{equation}
\end{small}
where $u \in [0, d_1-1]$ and $v \in [0, d_2-1]$ represent the Hartley indices.

\begin{figure}[!t]
    \centering
    \includegraphics[width=\linewidth]{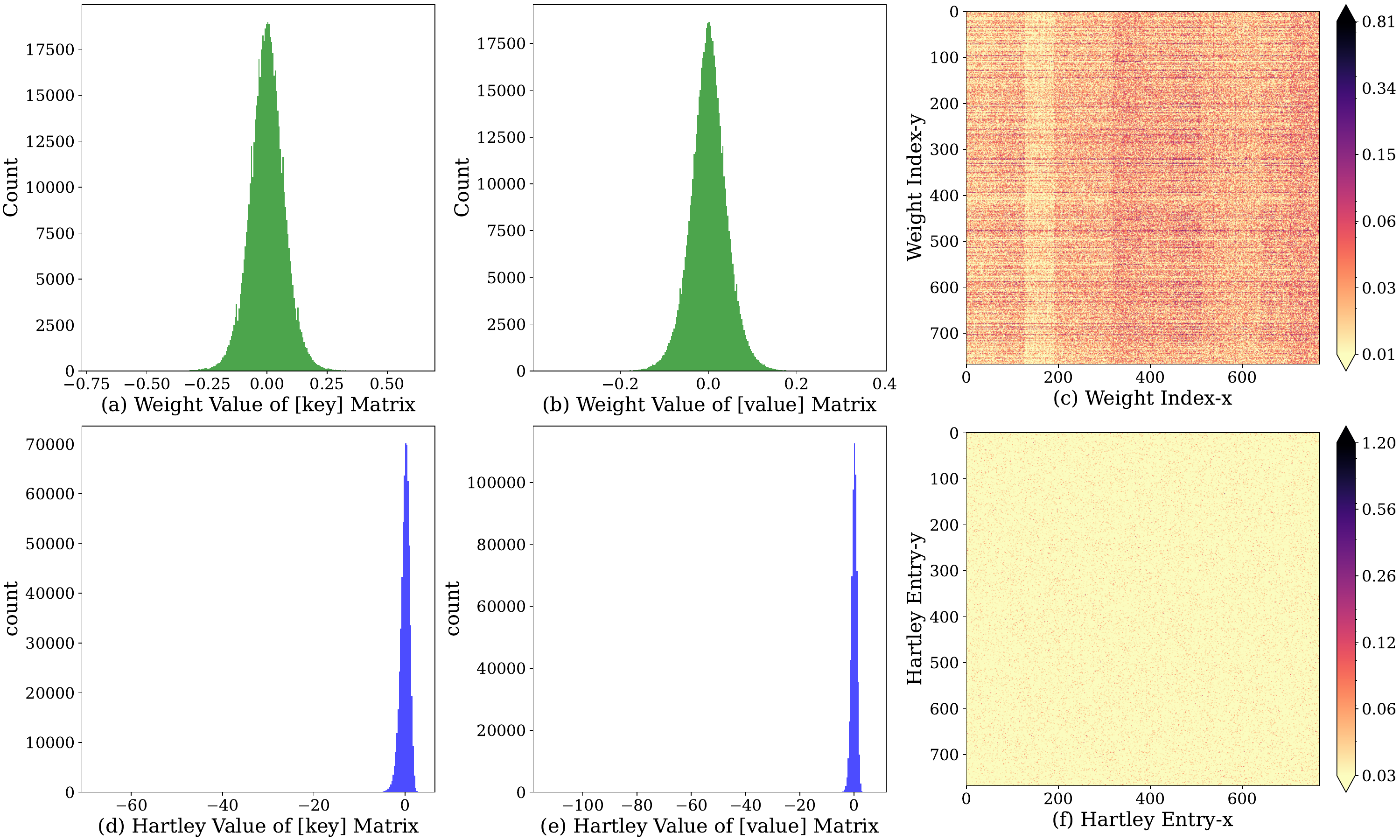}
    \caption{\small 
    Visual representation of the RoBERTa attention mechanism's key and value matrices before and after discrete Hartley transform (DHT). (a)(b) show the original weight distributions of the key and value matrices, respectively. (d)(e) depict the transformed DHT values, demonstrating effective spectral compression. Heatmaps (c)(f) illustrate the output weights before and after DHT, highlighting the achieved sparsity and efficient representation.}
    \label{fig:profile}
\end{figure}

As demonstrated in Fig.~\ref{fig:profile}, we analyze the key and value matrices of the RoBERTa-base model before and after applying the discrete Hartley transform, which effectively compresses the weight matrix into a compact spectral form. Similar trends are observed across models such as ViT, LLaMA, and VL-BART. To optimize parameter efficiency and reduce computational complexity, we selectively update only $n$ Hartley coefficients. This selection is driven by an energy-based method to capture the most informative coefficients, complemented by random initialization for diversity.

The energy of each Hartley component $E(u,v)$ is calculated as:

\begin{small}
\begin{equation}
    E(u,v) = \left| \mathbf{H}(u,v) \right|^2,
\end{equation}
\end{small}
where $\mathbf{H}(u,v)$ is the DHT coefficient at $(u,v)$. The top $n_{\text{energy}} = \lfloor \delta \cdot n \rfloor$ components with the highest energy are selected, while the rest are randomly chosen to ensure diversity.

\begin{small}
\begin{equation}
    \Delta \mathbf{H}_{\text{select}} = \Delta \mathbf{H}_{n_{\text{energy}}} + \Delta \mathbf{H}_{n_{\text{random}}}.
\end{equation}
\end{small}

After updating $\Delta \mathbf{H}$, the inverse DHT is applied to project the updates back into the spatial domain:

\begin{small}
\begin{equation}
\label{eq3}
    \mathbf{W} = \mathbf{W}_0 + \text{DHT}^{-1}(\Delta \mathbf{H}).
\end{equation}    
\end{small}

During the backward pass, the gradient with respect to the learnable Hartley coefficients is computed by projecting the spatial gradient $\frac{\partial \mathcal{L}}{\partial \mathbf{W}}$ into the spectral domain. To ensure that only the selected $n$ coefficients are updated, an indicator mask $\mathbf{M} \in \{0,1\}^{d_1 \times d_2}$ is employed to the gradient:

\begin{small}
\begin{equation}
    \frac{\partial \mathcal{L}}{\partial \Delta \mathbf{H}} = \mathbf{M} \circ \text{DHT} \left( \frac{\partial \mathcal{L}}{\partial \mathbf{W}} \right),
\end{equation}    
\end{small}
where $\circ$ denotes the element-wise multiplication. 

Finally, the selected $n$ Hartley coefficients are updated using the gradient descent:

\begin{small}
\begin{equation}
    \Delta \mathbf{H} \leftarrow \Delta \mathbf{H} - \eta \left( \mathbf{M} \circ \frac{\partial \mathcal{L}}{\partial \Delta \mathbf{H}} \right),
\end{equation}
\end{small}
where $\eta$ is the learning rate. This ensures that only the learnable Hartley coefficients are modified, maintaining sparsity.

\subsection{Parameter Efficiency Analysis}

\begin{table}[!t]
\centering
\resizebox{0.47\textwidth}{!}{%
\begin{tabular}{l|ccc|cccc}
\toprule
\textbf{Base Models} & \multicolumn{3}{c|}{\textbf{LoRA}} & \multicolumn{3}{c}{\textbf{SSH}} \\
\cmidrule(lr){2-4} \cmidrule(lr){5-7}
 & \textbf{r} & \textbf{\# Tr. Para.} & \textbf{Req. Bytes} & \textbf{n} & \textbf{\# Tr. Para.} & \textbf{Req. Bytes} \\
\midrule
\multirow{2}{*}{\textbf{RoBERTa Base}} & 4 & 147K & 574KB & 200 & 4.8K & 18.8KB \\
 & 8 & 295K & 1.13MB & 200 & 24K & 94KB \\
\midrule
\multirow{2}{*}{\textbf{RoBERTa Large}} & 4 & 393K & 1.5MB & 200 & 9.6K & 36.5KB \\
 & 8 & 786K & 3MB & 750 & 36.0K & 131.6KB \\
\midrule
\multirow{2}{*}{\textbf{GPT-2 Medium}} & 4 & 400K & 1.34MB & 375 & 18.1K & 65.8KB \\
 & 8 & 786K & 3MB & 750 & 36.0K & 131.6KB\\
\midrule
\multirow{2}{*}{\textbf{GPT-2 Large}} & 4 & 737K & 2.81MB & 375 & 18.1K & 105.8KB\\
 & 8 & 1.47M & 5.74MB & 750 & 36.0K & 211.5KB \\
\midrule
\multirow{2}{*}{\textbf{LLaMA-2 7B}} & 16 & 8.39M & 32.8MB & 750 & 48.0K & 187KB \\
 & 64 & 33.5M & 131.1MB & 1500 & 96.0K & 375KB \\
\midrule
\multirow{2}{*}{\textbf{LLaMA-2 13B}} & 16 & 13.1M & 51.2MB & 750 & 60K & 234KB \\
 & 64 & 52.4M & 204.8MB & 1500 & 120K & 469KB \\
 \midrule
\multirow{2}{*}{\textbf{LLaMA-3.1 8B}} & 16 & 13.1M & 51.2MB & 750 & 53.7K & 209KB \\
 & 64 & 52.4M & 204.8MB & 1500 & 107.5K & 420.1KB \\
\midrule
\multirow{2}{*}{\textbf{ViT Base}} & 8 & 295K & 1.13MB & 2250 & 54K & 210.7KB \\
 & 16 & 590K & 2.25MB & 7500 & 179.2K & 700.5KB \\
\midrule
\multirow{2}{*}{\textbf{ViT Large}} & 8 & 786K & 2.93MB & 2250 & 108K & 422.3KB \\
 & 16 & 1.57M & 6MB & 7500 & 350K & 1.38MB \\
\bottomrule
\end{tabular}%
}
\caption{Comparison of trainable parameters and required bytes between LoRA and SSH on different base models. SSH offers a substantial reduction in both trainable parameters and memory usage.}
\label{tab:nlu}
\end{table}

We evaluate the parameter efficiency and memory requirement of SSH in comparison to LoRA across several base models. The number of trainable parameters in SSH is given by \( |\Theta| = n \times L \), where \( n \) represents the number of selected frequencies, and \( L \) is the number of layers being fine-tuned. 
For LoRA, the parameter count is calculated as \( |\Theta| = r \times (d_1 + d_2) \times L \), where \( d_1 \) and \( d_2 \) are the dimensions of each layer, and \( r \) is the rank used for LoRA's low-rank decomposition.

Tab.~\ref{tab:nlu} compares SSH and LoRA in terms of trainable parameters and memory usage. SSH consistently requires fewer parameters and significantly less memory than LoRA. For instance, in the RoBERTa Base model, SSH with \( n = 200 \) uses only 4.8K parameters and 18.8KB of memory, while LoRA with \( r = 4 \) requires 147K parameters and 574KB of memory. This trend continues with larger models, such as LLaMA-2 13B, where SSH uses 60K parameters compared to LoRA's 13.1M parameters.

The memory efficiency of SSH becomes even more pronounced in larger models like ViT Large. For \( n = 2250 \), SSH requires just 108K parameters (422.3KB), while LoRA with \( r = 8 \) needs 786K parameters (2.93MB). This considerable reduction in both the parameter number and memory footprint highlights the scalability and efficiency of SSH, making it especially suitable for resource-constrained environments.

\section{Experiments}
SSH is compared against state-of-the-art parameter-efficient fine-tuning (PEFT) methods. The experiments are conducted across multiple domains, including single-modality tasks such as natural language understanding (NLU) and natural language generation (NLG), as well as instruction tuning, text summarization, and mathematical reasoning. Additionally, SSH is evaluated on multi-modality tasks, such as vision-language image classification. Finally, an ablation study is performed to assess the effectiveness of our approach.

\subsection{Baselines}
We compare SSH with the following baselines:
\begin{itemize}
    \item \textbf{Full Fine-Tuning (FF):} The entire model is fine-tuned, with updates to all parameters.
    \item \textbf{Adapter Tuning~\cite{houlsby2019parameter,lin2020exploring,ruckle2020adapterdrop,pfeiffer2020adapterfusion}:} Methods that introduce adapter layers between the self-attention and MLP modules for parameter-efficient tuning.
    \item \textbf{LoRA~\cite{hu2022lora}:} A method that estimates weight updates via low-rank matrices.
    \item \textbf{AdaLoRA~\cite{zhang2303adaptive}:} An extension of LoRA that dynamically reallocates the parameter budget based on importance scores.
    \item \textbf{DoRA~\cite{liu2024dora}:} Decomposes pretrained weights into magnitude and direction, using LoRA for efficient directional updates.
    \item \textbf{VeRA~\cite{kopiczko2023vera}:} Employs a single pair of low-rank matrices across all layers, to reduce parameters.
    \item \textbf{FourierFT~\cite{gao2024parameter}:} Fine-tunes models by learning a subset of spectral coefficients in the Fourier domain.
    \item 
    \textbf{AFLoRA~\cite{liu2024aflora}:} Freezes low-rank adaptation parameters using a learned freezing score, reducing trainable parameters while maintaining performance.
    \item 
    \textbf{LaMDA~\cite{azizi2024lamda}:} Fine-tunes large models via spectrally decomposed low-dimensional adaptation, reducing trainable parameters and memory usage while maintaining performance.
    
\end{itemize}

\subsection{Natural Language Understanding}

\begin{table*}[!ht]
\centering
\resizebox{0.85\textwidth}{!}{%
\begin{tabular}{cl|r|ccccccccc}
\toprule
& \textbf{Model} & \textbf{\# Trainable} & \textbf{SST-2} & \textbf{MRPC} & \textbf{CoLA} & \textbf{QNLI} & \textbf{RTE} & \textbf{STS-B} & \multirow{2}{*}{\textbf{Avg.}} \\
& \textbf{\& Method} & \textbf{Parameters} & \textbf{(Acc.)} & \textbf{(Acc.)} & \textbf{(MCC)} & \textbf{(Acc.)} & \textbf{(Acc.)} & \textbf{(PCC)} \\
\midrule
\multirow{9}{*}{\rotatebox{90}{\textbf{BASE}}} 
& FF & 125M & 94.8 & 90.2 & 63.6 & 92.8 & 78.7 & 91.2 & 85.22 \\
& BitFit & 0.1M & 93.7 & \textbf{92.7} & 62.0 & 91.8 & \textbf{81.5} & 90.8 & 85.42 \\
& Adpt\textsuperscript{D} & 0.9M & 94.7 & 88.4 & 62.6 & 93.0 & 75.9 & 90.3 & 84.15 \\
& LoRA & 0.3M & \textbf{95.1} & 89.7 & 63.4 & \textbf{93.3} & 78.4 & \textbf{91.5} & 85.23 \\
& AdaLoRA & 0.3M & 94.5 & 88.7 & 62.0 & 93.1 & 81.0 & 90.5 & 84.97 \\
& DoRA & 0.3M & 94.9 & 89.9 & 63.7 & \textbf{93.3} & 78.9 & \textbf{91.5} & 85.37 \\
& AFLoRA & 0.27M & 94.1 & 89.3 & 63.5 & 91.3 & 77.2 & 90.6 & 84.33 \\
& LaMDA & 0.06M &  94.6 & 89.7 & 64.9 & 91.7 & 78.2 & 90.4 & 84.92 \\
& VeRA & 0.043M & 94.6 & 89.5 & \textbf{65.6} & 91.8 & 78.7 & 90.7 & 85.15 \\
& FourierFT & 0.024M & 94.2 & 90.0 & 63.8 & 92.2 & 79.1 & 90.8 & 85.02 \\
\rowcolor{green!17}
& \textbf{SSH} & \textbf{0.018M} & 94.1 & 91.2 & 63.6 & 92.4 & 80.5 & 90.9 & \textbf{85.46} \\
\midrule
\multirow{8}{*}{\rotatebox{90}{\textbf{LARGE}}} 
& FF & 356M & 96.3 & 90.9 & 68.0 & 94.7 & 86.6 & 92.4 & 88.11 \\
& Adpt\textsuperscript{P} & 3M & 96.1 & 90.2 & \textbf{68.3} & 94.7 & 83.8 & 92.1 & 87.55 \\
& Adpt\textsuperscript{P} & 0.8M & \textbf{96.6} & 89.7 & 67.8 & 94.7 & 80.1 & 91.9 & 86.82 \\
& Adpt\textsuperscript{H} & 6M & 96.2 & 88.7 & 66.5 & 94.7 & 83.4 & 91.0 & 86.75 \\
& Adpt\textsuperscript{H} & 0.8M & 96.3 & 87.7 & 66.3 & 94.7 & 72.9 & 91.5 & 84.90 \\
& LoRA & 0.8M & 96.2 & 90.2 & 68.2 & \textbf{94.8} & 85.2 & 92.3 & 87.82 \\
& DoRA & 0.9M & 96.4 & \textbf{91.0} & 67.2 & \textbf{94.8} & 85.4 & 92.1 & 87.82 \\
& AFLoRA & 0.76M & 96.3 & 90.0 & 67.5 & 94.3 & 86.6 & 91.9 & 87.77 \\
& LaMDA & 0.093M &  96.2 & 90.1 & 68.1 & 94.5 & 87.3 & 92.0 & 88.03 \\
& VeRA & 0.061M & 96.1 & 90.9 & 68.0 & 94.4 & 85.9 & 91.7 & 87.83 \\
& FourierFT & 0.048M & 96.0 & 90.9 & 67.1 & 94.4 & \textbf{87.4} & 91.9 & 87.95 \\
\rowcolor{green!17}
& \textbf{SSH} & \textbf{0.036M} & 96.2 & 90.9 & 67.9 & 94.5 & \textbf{87.4} & \textbf{92.2} & \textbf{88.17} \\
\bottomrule
\end{tabular}%
}
\caption{\small Performance of various fine-tuning methods on GLUE benchmark, using base and large models. Metrics include MCC for CoLA, PCC for STS-B, and accuracy for other tasks. Results are medians of 5 runs with different seeds; the best scores in each category are bolded. SSH delivers the best average performance across tasks while using significantly fewer trainable parameters.}
\label{tab:nlup}
\end{table*}

\noindent \textbf{Models and Datasets.}  
We evaluate SSH on the GLUE benchmark~\cite{wang2019glue} using RoBERTa~\cite{liu2019roberta} in both Base and Large configurations. The GLUE benchmark comprises a diverse set of NLU tasks, offering a comprehensive evaluation framework.

\noindent \textbf{Implementation Details.}  
The SSH method uses 750 of the 768\textsuperscript{2} available spectral coefficients for RoBERTa Base and 1024\textsuperscript{2} for RoBERTa Large, ensuring that each layer retains the most important spectral components. This selection remains consistent across all layers. To ensure fair comparison, we follow the same experimental settings as LoRA and FourierFT. Additional hyperparameters and details are provided in Tab.~\ref{tab:nluh} in the appendix~\ref{gluebench}.

\noindent \textbf{Results and Analysis} 
The results in Table~\ref{tab:nlup} indicate that SSH consistently delivers competitive performance across diverse NLU tasks while maintaining a significantly lower number of trainable parameters. Notably, SSH achieves 80.5\% accuracy on RTE, 92.4\% on QNLI, and 90.9 on STS-B, demonstrating its capability to generalize effectively across multiple linguistic tasks.

SSH also maintains robust performance in sentiment classification, achieving 94.1\% accuracy on SST-2, which is on par with other parameter-efficient methods such as LoRA and BitFit. On CoLA, SSH attains a score of 63.6, matching FourierFT and outperforming Adpt\textsuperscript{D} and AdaLoRA. Additionally, SSH exhibits strong generalization on MRPC with 91.2\% accuracy and achieves a 90.9 Pearson correlation on STS-B, further reinforcing its effectiveness across textual similarity and entailment tasks. These findings highlight SSH as a highly efficient and scalable fine-tuning approach, capable of achieving state-of-the-art performance with minimal parameter overhead.


\begin{table}[!t]
\centering
\scalebox{0.63}{
\begin{tabular}{l|lr|crcccccc}
\toprule
 & \textbf{Method} & \textbf{\# Tr. Para.} & \textbf{BLEU} & \textbf{NIST} & \textbf{METE.} & \textbf{ROU-L} & \textbf{CIDEr} \\
\midrule
\multirow{9}{*}{\rotatebox{90}{\textbf{GPT-2 Medium}}} 
& FT\textsuperscript{1} & 354.92M & 68.2 & 8.62 & 46.2 & 71.0 & 2.47 \\
& Adpt\textsuperscript{L\textsuperscript{1}} & 0.37M & 66.3 & 8.41 & 45.0 & 69.8 & 2.40 \\
& Adpt\textsuperscript{L\textsuperscript{1}} & 11.09M & 68.9 & 8.71 & 46.1 & 71.3 & 2.47 \\
& Adpt\textsuperscript{H\textsuperscript{1}} & 11.09M & 67.3 & 8.50 & 46.0 & 70.7 & 2.44 \\
& LoRA & 0.35M & 68.9 & 8.76 & 46.6 & 71.5 & 2.51 \\
& DoRA & 0.36M & 69.2 & 8.79 & 46.9 & 71.7 & 2.52\\
& VeRA & 0.35M & \textbf{70.1} & 8.81 & 46.6 & 71.5 & 2.50 \\
& FourierFT & 0.048M & 69.1 & \textbf{8.82} & 47.0 & 71.8 & 2.51 \\
\rowcolor{green!17}
& \textbf{SSH} & \textbf{0.036M} & \textbf{70.1} & \textbf{8.82} & \textbf{47.2} & \textbf{71.9} & \textbf{2.54} \\
\midrule
\multirow{8}{*}{\rotatebox{90}{\textbf{GPT-2 Large}}} 
& FT\textsuperscript{1} & 774.03M & 68.5 & 8.78 & 46.0 & 69.9 & 2.45 \\
& Adpt\textsuperscript{L\textsuperscript{1}} & 0.88M & 69.1 & 8.68 & 46.1 & 71.0 & 2.49 \\
& Adpt\textsuperscript{L\textsuperscript{1}} & 23.00M & 68.9 & 8.70 & 46.1 & 71.3 & 2.45 \\
& LoRA & 0.77M & 69.4 & 8.81 & 46.5 & \textbf{71.9} & 2.50 \\
& DoRA & 0.79M & 69.8 & 8.83 & 46.9 & \textbf{71.9} & 2.50 \\
& VeRA & 0.17M & \textbf{70.3} & 8.85 & 46.6 & 71.6 & 2.54 \\
& FourierFT & 0.072M & 70.2 & 8.90 & 47.0 & 71.8 & 2.50 \\
\rowcolor{green!17}
& \textbf{SSH} & \textbf{0.054M} & \textbf{70.3} & \textbf{8.93} & \textbf{47.2} & \textbf{71.9} & \textbf{2.55} \\
\bottomrule
\end{tabular}}
\caption{\small Performance comparison of various fine-tuning methods on GPT-2 Medium and Large models, evaluated using BLEU, NIST, METEOR, ROUGE-L, and CIDEr metrics. \textsuperscript{1} denotes results sourced from previous studies. }
\label{tab:e2e}
\end{table}

\subsection{Natural Language Generation}

\noindent \textbf{Models and Datasets.}  
We evaluate SSH on the E2E natural language generation (NLG) task~\cite{novikova2017e2e}, fine-tuning GPT-2 Medium and Large models~\cite{radford2019language}, which consist of 24 and 36 transformer blocks.

\noindent \textbf{Implementation Details.}  
We fine-tune LoRA, DoRA, FourierFT, VeRA, and the proposed SSH on GPT-2 Medium and Large, using a linear learning rate scheduler over 5 epochs. Results are averaged across 3 runs, with detailed hyperparameters provided in Tab.~\ref{tab:nlgh} in the Appendix~\ref{gluebench}.

\noindent \textbf{Results and Analysis.}  
As shown in Tab.~\ref{tab:e2e}, SSH consistently delivers superior or comparable performance across all evaluation metrics, while requiring significantly fewer trainable parameters. For GPT-2 Medium, SSH matches the highest BLEU score (70.1) and outperforms other methods in NIST (8.82), METEOR (47.2), ROUGE-L (71.9), and CIDEr (2.54), all with 10.3\% fewer parameters than LoRA and 25\% fewer than FourierFT. A similar trend is observed for GPT-2 Large, where SSH achieves the highest NIST (8.93) and METEOR (47.2) scores, while maintaining a 7.1\% parameter reduction compared to LoRA.

\begin{table}[!t]
\centering
\resizebox{0.5\textwidth}{!}{%
\begin{tabular}{l|l|c|crcc}
\toprule
\textbf{Model} & \textbf{Method} & \textbf{\# Tr. Para.} & \textbf{MT-Bench} & \textbf{Vicuna} \\
\midrule
\multirow{5}{*}{\textbf{LLaMA2-7B}} 
& LoRA & 159.9M & 5.19 & 7.37 \\
& DoRA & 163.7M & 5.20 & 7.41 \\
& VeRA & 1.6M & 5.18 & 7.47 \\
& FourierFT & 0.064M & 5.09 & 7.50 \\
\rowcolor{green!17}
& \textbf{SSH} & \textbf{0.048M} & \textbf{5.22} & \textbf{7.51}\\
\midrule
\multirow{5}{*}{\textbf{LLaMA2-13B}} 
& LoRA & 250.3M & 5.77 & 7.89\\
& DoRA & 264.5M & 5.79 & 7.90 \\
& VeRA & 2.4M & \textbf{5.93} & 7.90 \\
& FourierFT & 0.08M & 5.82 & 7.92 \\
\rowcolor{green!17}
& \textbf{SSH} & \textbf{0.06M} & \textbf{5.93} & \textbf{7.95} \\
\midrule
\multirow{5}{*}{\textbf{LLaMA3.1-8B}} 
& LoRA & 183.3M & 5.65 & 7.52 \\
& DoRA & 186.9M & 5.66 & \textbf{7.59} \\
& VeRA & 1.9M & 5.61 & 7.49 \\
& FourierFT & 0.073M & 5.67 & 7.67 \\
\rowcolor{green!17}
& \textbf{SSH} & \textbf{0.055M} & \textbf{5.69} & \textbf{7.71} \\
\bottomrule
\end{tabular}%
}
\caption{\small Performance comparison of fine-tuning methods on LLaMA models using the Alpaca dataset. Evaluation scores on MT-Bench and Vicuna are generated and scored by GPT-4.}
\label{tab:mtbench_vicuna}
\end{table}

\subsection{Instruction Tuning}

\noindent \textbf{Models and Datasets.}  
We fine-tune LLaMA2-7B, LLaMA2-13B, and LLaMA3.1-8B using SSH and baseline methods on the Alpaca dataset~\cite{taori2023stanford}. For evaluation, we generate responses to predefined questions from the MT-Bench~\cite{zheng2024judging} and Vicuna Eval datasets, which are then scored by GPT-4 on a 10-point scale.

\noindent \textbf{Implementation Details.}  
Following previous work~\cite{dettmers2024qlora,dettmers20228bit}, we apply LoRA, DoRA, and VeRA to all linear layers except the top one. For FourierFT, we use the configuration from~\cite{gao2024parameter}, and for SSH, we set \(n = 750\). All models are trained using QLoRA’s quantization technique~\cite{dettmers2024qlora} on a single GPU. Each method is trained for one epoch, and we report the average score across all generated responses. Hyperparameter details are provided in Tab.\ref{tab:hyperparamsIn} in the Appendix~\ref{gluebench}.

\noindent \textbf{Results and Analysis.}  
The results in Tab.~\ref{tab:mtbench_vicuna} clearly demonstrate the significant efficiency of SSH compared to other fine-tuning methods such as LoRA, DoRA, and FourierFT. For LLaMA2-7B, SSH achieves the best MT-Bench (5.22) and Vicuna (7.51) scores while reducing trainable parameters by over 99.7\%, using only 0.048M parameters compared to LoRA's 159.9M. Similarly, in LLaMA2-13B, SSH ties with VeRA for the highest MT-Bench score (5.93) and surpasses all methods in Vicuna (7.95), again achieving this with a drastically lower parameter count (0.06M vs. 250.3M for LoRA). Even in the larger LLaMA3.1-8B model, SSH continues to outperform, leading in MT-Bench (5.69) and maintaining a competitive Vicuna score (7.71) with far fewer parameters (0.055M).

\begin{table}
\centering
\resizebox{0.47\textwidth}{!}{%
\begin{tabular}{l|l|r|cccccc}
\toprule
\textbf{Model} & \textbf{Method} & \textbf{\# Train. Para.} & \textbf{CIFAR100} & \textbf{DTD} & \textbf{EuroSAT} & \textbf{OxfordPets} \\
\midrule
\multirow{7}{*}{\textbf{ViT-B}} 
& Head & - & 84.3 & 69.8 & 88.7 & 90.3 \\
& Full & 85.8M & \textbf{92.4} & \textbf{77.7} & \textbf{99.1}& \textbf{93.4} \\
& LoRA & 581K & 92.1 & 75.2 & 98.4 & 93.2 \\
& Dora & 594K & 92.3 & 75.3 & 98.7 & 93.2 \\
& VeRA & 57.3K & 91.7 & 74.6 & 98.5 & \textbf{93.4}\\
& FourierFT & 72K & 94.2 & 75.1 & 98.8 & 93.2 \\
\rowcolor{green!17}
& \textbf{SSH} & \textbf{54K} & 91.6 & 76.1 & \textbf{99.1} & \textbf{93.4} \\

\midrule
\multirow{7}{*}{\textbf{ViT-L}} 
& Head & - & 84.7 & 73.3 & 92.6 & 91.1 \\
& Full & 303.3M & 93.6 & 81.8 & \textbf{99.1} & 94.4 \\
& LoRA & 1.57M &  94.9 & 81.8 & 98.63 & \textbf{94.8} \\
& Dora & 1.62M &  \textbf{95.1} & 81.8 &   98.8 & \textbf{94.8}\\
& VeRA & 130.5K & 94.2 & 81.6& 98.6 & 93.7 \\
& FourierFT & 144K & 93.7 & 81.2 & 98.7 & 94.5 \\
\rowcolor{green!17}
& \textbf{SSH} & \textbf{108K} & 94.5 & \textbf{81.9} & 99.0& \textbf{94.8} \\
\bottomrule
\end{tabular}%
}
\caption{\small Performance of various fine-tuning methods on ViT-B and ViT-L models across different datasets. The best results for each dataset are highlighted in bold. The best results are highlighted in bold. SSH offers strong parameter efficiency, excelling on DTD and EuroSAT while delivering competitive performance on CIFAR100 and OxfordPets, making it a balanced solution for various vision tasks.}
\label{tab:vit_results}
\end{table}

\subsection{Text Summarization}
\noindent \textbf{Models and Datasets.}  
We evaluate the effectiveness of SSH against other baseline methods on the BART-Large model~\cite{lewis2019bart} for text summarization tasks. Specifically, we assess its performance on the XSUM~\cite{narayan2018don} and CNN/DailyMail~\cite{hermann2015teaching} datasets.

\begin{table}[!t]
    \centering
    \resizebox{0.52\textwidth}{!}{%
    \begin{tabular}{l|c|c|c}
        \toprule
        \textbf{Method} & \textbf{Para. (M)} & \textbf{XSUM} & \textbf{CNN/DailyMail} \\
        \midrule
        AFLoRA ($r=32$) & 5.27 & 44.71/21.92/37.33 & 44.95/21.87/42.25 \\
        LaMDA ($r=32$) & 0.85 & 43.94/20.69/35.21 & 44.16/21.17/40.48 \\
        \rowcolor{green!17}
        SSH ($n=5000$) & 0.21 & 44.72/22.05/37.42 & 44.89/21.75/42.13 \\
        \bottomrule
    \end{tabular}}
    \caption{Performance comparison of SSH, AFLoRA, and LaMDA on BART-Large for text summarization tasks. Results are reported as ROUGE-1/ROUGE-2/ROUGE-L.}
    \label{tab:nlg_bart}
\end{table}

\noindent \textbf{Implementation Details.}  
We compare SSH against AFLoRA and LaMDA under consistent experimental conditions. For AFLoRA and LaMDA, we set the rank $r=32$, while for SSH, we select $n=5000$ Hartley spectrum points. The models are trained using a learning rate of $2\times10^{-4}$, with a batch size of 32 for XSUM and 64 for CNN/DailyMail. Training is conducted for 25 epochs on XSUM and 15 epochs on CNN/DailyMail.

\noindent \textbf{Results and Analysis.}  
Table~\ref{tab:nlg_bart} presents the ROUGE evaluation scores (ROUGE-1/ROUGE-2/ROUGE-L) for different fine-tuning approaches. SSH achieves competitive performance while utilizing significantly fewer trainable parameters compared to AFLoRA and LaMDA. On the XSUM dataset, SSH attains the highest ROUGE-2 score (22.05), surpassing AFLoRA (21.92) and LaMDA (20.69) by 0.13 and 1.36 points, respectively. Furthermore, SSH achieves the highest ROUGE-L score (37.42), outperforming AFLoRA by 0.09 and LaMDA by 2.21 points.

Similarly, on the CNN/DailyMail dataset, SSH attains a ROUGE-1 score of 44.89, which is marginally lower than AFLoRA (44.95) by 0.06 points, but it outperforms LaMDA (44.16) by 0.73 points. In terms of ROUGE-2, SSH achieves 21.75, trailing AFLoRA (21.87) by 0.12 points but exceeding LaMDA (21.17) by 0.58 points. Additionally, SSH attains a ROUGE-L score of 42.13, which is 0.12 points lower than AFLoRA but significantly higher than LaMDA by 1.65 points. Overall, SSH consistently demonstrates strong performance while requiring significantly fewer trainable parameters (0.21M) compared to AFLoRA (5.27M) and LaMDA (0.85M). 

\subsection{Mathematical Reasoning}

\noindent \textbf{Models and Dataset.} We evaluate the performance of SSH against AFLoRA and LaMDA on the LLaMA3.1-8B model using the GSM8K~\cite{cobbe2021training}, a widely used dataset designed to assess mathematical reasoning abilities.

\noindent \textbf{Implementation Details.} All methods are trained with a learning rate of $3\times10^{-4}$ for six epochs using a batch size of 16. For parameter-efficient fine-tuning, AFLoRA and LaMDA employ a low-rank adaptation setting of $r=32$, while SSH leverages a Hartley spectrum selection with $n=10000$. Table~\ref{tab:llama_gsm8k} presents a comparison of the methods in terms of trainable parameters and accuracy on GSM8K.

\begin{table}[!t]
    \centering
    \resizebox{0.5\textwidth}{!}{%
    \begin{tabular}{l|c|c}
        \toprule
        \textbf{Method} & \textbf{Trainable Parameters (M)} & \textbf{GSM8K Accuracy} \\
        \midrule
        AFLoRA ($r=32$) & 20.23 & 38.63 \\
        LaMDA ($r=32$) & 4.99 & 38.11 \\
        \rowcolor{green!17}
        SSH ($n=10000$) & 1.54 & \textbf{38.67} \\
        \bottomrule
    \end{tabular}}
    \caption{Comparison of SSH with AFLoRA and LaMDA on LLaMA3.1-8B for GSM8K. Accuracy is reported as a percentage.}
    \label{tab:llama_gsm8k}
\end{table}

\noindent \textbf{Results and Analysis.} SSH achieves the highest accuracy (38.67\%), surpassing AFLoRA (38.63\%) and LaMDA (38.11\%) while using significantly fewer trainable parameters. SSH requires only 1.54M parameters, representing a \textbf{92.4\% reduction} compared to AFLoRA and a \textbf{69.1\% reduction} compared to LaMDA. 

Despite having nearly 13 times fewer parameters than AFLoRA, SSH achieves comparable accuracy, demonstrating a superior trade-off between efficiency and performance. While LaMDA exhibits the lowest accuracy, SSH maintains robustness in mathematical reasoning tasks with minimal resource requirements.


\subsection{Image Classification}

\noindent \textbf{Models and Datasets.}  
We evaluate our method on the Vision Transformer (ViT)~\cite{dosovitskiy2020image}, using both the Base and Large variants. Image classification is performed on the CIFAR-100~\cite{krause20133d}, DTD~\cite{cimpoi2014describing}, EuroSAT~\cite{helber2019eurosat}, and OxfordPets~\cite{parkhi2012cats} datasets.

\noindent \textbf{Implementation Details.}  
We evaluate SSH, LoRA, DoRA, VeRA, and FourierFT by applying them to the query and value layers of ViT. Training only the classification head is denoted as "Head". We set \( r = 16 \) for LoRA, \( n = 3000 \) for FourierFT, and \( n = 2250 \) for SSH. Learning rates and weight decay are tuned for all methods, with training limited to 10 epochs. Further hyperparameter details are provided in Tab.~\ref{tab:SSH_image} in the Appendix~\ref{gluebench}.

\noindent \textbf{Results and Analysis.}  
Tab.~\ref{tab:vit_results} highlights the performance of various fine-tuning methods on ViT-B and ViT-L across four image classification datasets. For the ViT-B model, SSH delivers competitive results with only 54K trainable parameters, significantly fewer than LoRA and DoRA, which use more than 10 times as many. Notably, SSH matches the full fine-tuning performance on EuroSAT and OxfordPets, achieving 99.1\% and 93.4\% accuracy, respectively. For the ViT-L model, SSH also proves efficient, achieving near-optimal performance with only 108K parameters. It sets the highest score on DTD with 81.9\% accuracy and matches the best performance on OxfordPets at 94.8\%.

\subsection{Ablation Study}
\begin{figure}
    \centering
    \includegraphics[width=\linewidth]{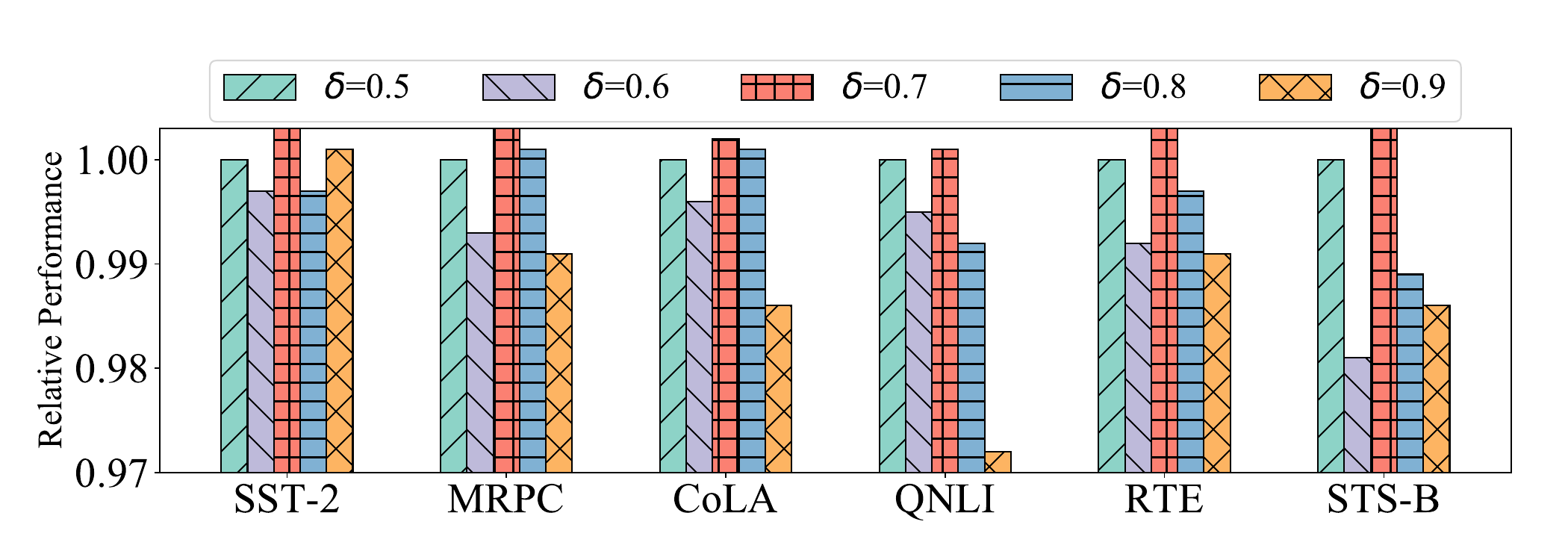}
    \caption{\small Ablation study of SSH on GLUE tasks illustrating the effect of varying energy ratios ($\delta$) on performance with RoBERTa-base (n=750). Performance is normalized to $\delta = 0.5$, showing optimal balance and diversity in spectral representation at $\delta = 0.7$.
}
    \label{fig:ratio}
\end{figure}

\noindent \textbf{Energy Ratio Ablation Study.}
\label{subsubsec:energyratio}
Figure~\ref{fig:ratio} presents an ablation study of SSH across GLUE tasks with varying energy ratios (\(\delta\)) on RoBERTa-base with \(n=750\), where performance is normalized to \(\delta = 0.5\). The energy ratios considered are \(\delta=0.5\), \(\delta=0.6\), \(\delta=0.7\), \(\delta=0.8\), and \(\delta=0.9\). 


The ablation study indicates that an energy ratio of \(\delta=0.7\) optimally balances the selection of spectral components, consistently enhancing performance across natural language understanding tasks such as MRPC and CoLA. This balance prevents overfitting and underfitting, ensuring the retention of informative frequencies while excluding those that are redundant. In contrast, lower ratios (\(\delta=0.5\) or \(\delta=0.6\)) result in inadequate frequency representation, adversely affecting performance in tasks that require robust syntactic and semantic analysis, such as QNLI and CoLA. Higher ratios (\(\delta=0.8\) and \(\delta=0.9\)), while expanding the range of considered frequencies, often introduce noise that compromises the model's focus and generalization ability, particularly evident in tasks like QNLI and STS-B.


\section{Limitations}

While SSH shows great promise, there are several limitations that need to be addressed in the future:

\begin{itemize}
    \item \textbf{One-go Additional Computational overhead:} A notable limitation is the need to perform a one-go Discrete Hartley Transformation (DHT) on pre-trained weights to guide the selection of the most informative frequencies. This step introduces additional computational overhead and memory requirements upfront.

    \item \textbf{Generalization across Different Tasks:} Although SSH has demonstrated strong performance across various tasks, its effectiveness might vary depending on the task or the structure of the data. Due to specific model characteristics, certain tasks may favor other methods, such as LoRA or FourierFT.


    \item \textbf{Dependency on Pretrained Model Quality:} Since SSH is a fine-tuning method, its performance is inherently dependent on the quality of the pretrained model. If the pretrained model is suboptimal, SSH may not yield significant improvements over other methods.

\end{itemize}

\section{Conclusion}

We introduced Sparse Spectrum Adaptation via Discrete Hartley Transformation (SSH), a novel PEFT method that reduces the number of trainable parameters while maintaining competitive performance. SSH leverages the real-valued DHT and its symmetric forward and backward transform to selectively update the most informative spectral components, addressing the computational and memory challenges of fine-tuning large models. Through extensive experiments across diverse tasks, SSH demonstrates robust versatility, excelling in single-modality NLP tasks such as natural language understanding (NLU), natural language generation (NLG), text summarization, and mathematical reasoning. Furthermore, SSH extends its effectiveness to multi-modality applications, including vision-language image classification. SSH not only achieves state-of-the-art performance but also surpasses existing PEFT methods in both parameter and computational efficiency, positioning it as a scalable and lightweight solution for fine-tuning large models across various domains.

\bibliography{0_acl_main}
\clearpage
\label{sec:appendix}
\section{Appendix}
\label{appendix}

In this supplementary material, we first provide detailed information about the datasets used in our experiments. Next, we outline the implementation specifics and hyper-parameter settings. We then present additional experimental results that further validate the effectiveness of the proposed SSH method. Finally, we include examples of instruction tuning to highlight the practical application of our approach SSH.

\subsection{Details of Datasets}
\paragraph{GLUE Benchmark.} 
\label{gluebench}
The General Language Understanding Evaluation (GLUE) benchmark\cite{wang2019glue} is a widely used platform for evaluating and advancing progress in natural language understanding (NLU). It includes nine tasks covering various NLU challenges, such as sentiment analysis, paraphrase detection, linguistic acceptability, natural language inference, and textual similarity. Notable tasks include the Stanford Sentiment Treebank (SST-2), which focuses on binary sentiment classification of movie reviews, and the Microsoft Research Paraphrase Corpus (MRPC), which assesses whether two sentences are semantically equivalent. The Corpus of Linguistic Acceptability (CoLA) measures a model's ability to distinguish grammatically correct sentences from incorrect ones, reflecting syntactic judgment skills. By offering diverse tasks and including some with limited training data, the benchmark promotes the development of models that generalize well across different language tasks and genres.

In addition to single-sentence classification tasks, GLUE includes several sentence-pair tasks. The Question Natural Language Inference (QNLI) task, derived from the Stanford Question Answering Dataset (SQuAD), requires models to determine whether a given context sentence contains the answer to a corresponding question. The Recognizing Textual Entailment (RTE) task combines several textual entailment datasets from diverse domains, such as news and Wikipedia, to test if a hypothesis can be logically inferred from a premise. The Semantic Textual Similarity Benchmark (STS-B) measures the similarity between sentence pairs using a regression-based approach, where models predict similarity scores on a continuous scale.

\paragraph{E2E Benchmark.}
The E2E dataset~\cite{novikova2017e2e} is designed for training and evaluating end-to-end data-driven natural language generation (NLG) systems within the restaurant domain. Comprising over 50,000 instances, it is notable for its linguistic complexity, including greater lexical diversity, syntactic variation, and discourse phenomena compared to earlier datasets. Evaluation is primarily based on five metrics: BLEU, NIST, METEOR, ROUGE-L, and CIDEr. BLEU measures n-gram overlap between the generated text and human references, emphasizing precision. METEOR accounts for synonymy and stemming, providing a more nuanced assessment of similarity. ROUGE-L evaluates fluency and structure through the longest common subsequence. CIDEr, by weighting n-grams according to their relevance in human references, offers a comprehensive measure of output quality.

\paragraph{Instruction Tuning Related Benchmarks}
The Alpaca dataset~\cite{taori2023stanford} consists of 51K instruction-following examples generated using OpenAI's text-davinci-003. It was developed to fine-tune Meta's LLaMA 7B model into a lightweight, instruction-following model called Alpaca. The dataset spans a wide range of tasks, including question-answering, summarization, and classification, allowing the fine-tuned model to perform similarly to much larger models while being more cost-efficient. A specific example is shown below:

\begin{tcolorbox}[colback=green!5!white, colframe=black!75!, sharp corners, boxrule=1pt]
\footnotesize
\texttt{
\{\\
\hspace*{0.5cm} "instructions": Translate the sentence from English to Spanish.\\
\hspace*{0.5cm} "input": The weather is beautiful today.\\
\hspace*{0.5cm} "output": El clima está hermoso hoy.\\
\}
}
\end{tcolorbox}

MT-Bench~\cite{zheng2024judging} is a recently introduced benchmark designed to evaluate the instruction-following capabilities of language foundation models. It features a set of open-ended questions aimed at assessing model performance across a variety of domains, including writing, roleplay, reasoning, mathematics, coding, information extraction, STEM, and the humanities. MT-Bench effectively distinguishes these abilities through domain-specific questions, providing a more comprehensive evaluation of model performance. An example from the benchmark is shown below.

\begin{tcolorbox}[colback=green!5!white, colframe=black!75!, sharp corners, boxrule=1pt]
\footnotesize
\texttt{
\{\\
\hspace*{0.5cm} "Q1": How many days are there in a leap year?\\
\hspace*{0.5cm} "Q2": How many days are there in two consecutive leap years?\\
\hspace*{0.5cm} "Solution": Q1: There are 366 days in a leap year. Q2: There are 732 days in two consecutive leap years.\\
\}
}
\end{tcolorbox}

Vicuna Eval~\cite{chiang2023vicuna} is a benchmark designed to evaluate the alignment of large language models (LLMs) with human preferences and is the predecessor to MT-Bench. Vicuna Eval assesses models across a wide range of topics, including coding, writing, mathematics, counterfactual reasoning, Fermi estimation, common sense, roleplay, knowledge, and general tasks. It offers a comprehensive framework for gauging how well models meet human expectations across various scenarios. An example from this evaluation is shown below.

\begin{tcolorbox}[colback=green!5!white, colframe=black!75!, sharp corners, boxrule=1pt]
\small
\texttt{
\{\\
\hspace*{0.5cm} "question": Describe the difference between supervised and unsupervised learning.\\
\hspace*{0.5cm} "category": machine learning.\\
\}
}
\end{tcolorbox}

\paragraph{Image Classification Datasets.}
Tab.~\ref{tab-image} provides detailed information about four widely-used vision datasets: CIFAR100, DTD, EuroSAT, and OxfordPets. It outlines key statistics, including the number of training (\#Train), validation (\#Val), and test (\#Test) samples, as well as the number of classes (\#Class) in each dataset. These datasets cover a range of domains, from object recognition (CIFAR100~\cite{krizhevsky2009learning}) and texture classification (DTD~\cite{cimpoi2014describing}) to satellite image classification (EuroSAT~\cite{helber2019eurosat}) and pet identification (OxfordPets~\cite{parkhi2012cats}). This diversity ensures that models are tested on various visual tasks, providing a robust evaluation of their performance.

To maintain consistency in model evaluation, all datasets are rescaled to a resolution of 224 × 224. This standardized input size simplifies comparisons by ensuring that all models receive uniformly sized images, which is essential for fair benchmarking. The datasets vary significantly in terms of size and complexity, with CIFAR100 containing the largest number of samples (60,000) across 100 classes, while OxfordPets is more specialized, focusing on 37 classes. This variety highlights the unique challenges posed by each dataset, contributing to comprehensive model assessments.

\begin{table}
\centering
\resizebox{0.47\textwidth}{!}{%
\begin{tabular}{l|r|r|r|r|l}
\toprule
\textbf{Dataset} & \textbf{\#Train} & \textbf{\#Val} & \textbf{\#Test} & \textbf{\#Class} & \textbf{Rescaled res.} \\
\midrule
CIFAR100  & 45,000 & 5,000 & 10,000 & 100 & \multirow{4}{*}{224 $\times$ 224} \\
DTD  & 4,060 & 452 & 1,128 & 47 &  \\
EuroSAT & 16,200 & 5,400 & 5,400 & 10 &  \\
OxfordPets & 3,312 & 368 & 3,669 & 37 & \\
\bottomrule
\end{tabular}%
}
\caption{Details about the vision datasets.}
\label{tab-image}
\end{table}

\subsection{Hyperparamaters}

\begin{table*}
\centering
\resizebox{0.7\textwidth}{!}{%
\begin{tabular}{ll|ccccccc}
\toprule
\textbf{Model} & \textbf{Hyperparameter} & \textbf{STS-B} & \textbf{RTE} & \textbf{MRPC} & \textbf{CoLA} & \textbf{SST-2} & \textbf{QNLI} \\
\midrule
\multirow{7}{*}{\textbf{Both}} 
& Optimizer & \multicolumn{6}{c}{AdamW} \\
& LR Schedule & \multicolumn{6}{c}{Linear} \\
& Warmup Ratio & \multicolumn{6}{c}{0.06} \\
& Frequency Bias & \multicolumn{6}{c}{False} \\
& $n_{SSH}$ & \multicolumn{6}{c}{750} \\
& $n_{FourierFT}$ & \multicolumn{6}{c}{1000} \\
\midrule
\multirow{7}{*}{\textbf{Base}} 
& Epochs & 60 & 90 & 30 & 100 & 40 & 40 \\
& Learning Rate (SSH) & 9E-2 & 9E-2 & 5E-2 & 1.2E-1 & 5E-2 & 1E-2 \\
& Learning Rate (FourierFT) & 9E-2 & 9E-2 & 5E-2 & 1.2E-1 & 5E-2 & 1E-2 \\
& Learning Rate (VeRA) & 9E-2 & 9E-2 & 5E-2 & 1.2E-1 & 5E-2 & 1E-2 \\
& Learning Rate (Head) & 9E-3 & 1.1E-2 & 6E-3 & 8E-3 & 6E-3 & 1E-3 \\
& Max Seq. Len & 512 & 512 & 512 & 512 & 512 & 512 \\
& Scaling Value & 84 & 110 & 141 & 49 & 140 & 29 \\
& Batch Size & 32 & 32 & 32 & 32 & 32 & 32 \\
\midrule
\multirow{7}{*}{\textbf{Large}} 
& Epochs & 30 & 60 & 30 & 80 & 10 & 30 \\
& Learning Rate (SSH) & 7E-2 & 8E-2 & 6E-2 & 4.3E-2 & 4.3E-2 & 6E-2 \\
& Learning Rate (FourierFT) & 7E-2 & 8E-2 & 6E-2 & 4.3E-2 & 4.3E-2 & 6E-2 \\
& Learning Rate (VeRA) & 7E-2 & 8E-2 & 6E-2 & 4.3E-2 & 4.3E-2 & 6E-2 \\
& Learning Rate (Head) & 1E-3 & 5E-3 & 1E-3 & 1.1E-2 & 1E-3 & 5E-3 \\
& Max Seq. Len & 512 & 512 & 512 & 256 & 128 & 512 \\
& Scaling Value & 121 & 90 & 120 & 90 & 69 & 69 \\
& Batch Size & 32 & 32 & 32 & 128 & 32 & 32 \\
\bottomrule
\end{tabular}%
}
\caption{\small Hyperparameters used for SSH across various GLUE tasks.}
\label{tab:nluh}
\end{table*}

\paragraph{Hyperparameters on GLUE Benchmarks.} Tab.~\ref{tab:nluh} summarizes the key hyperparameters used in experiments across various GLUE tasks for both Base and Large models. The table includes details on learning rate schedules, optimizer settings, warmup ratios, and seed values to ensure reproducibility. For both model sizes, the AdamW optimizer is employed with a linear learning rate schedule and a warmup ratio of 0.06. The frequency bias is set to false, and the frequency coefficient \( n \) is fixed at 750 for SSH unless otherwise specified. Each experiment is run using five different seeds: \{0, 11111, 22222, 33333, 44444\}.

For Base models, the number of training epochs varies between 30 and 100, depending on the task, with SST-2 requiring the longest training time. The FourierFT and SSH methods use a higher learning rate for base model training compared to the learning rate applied during fine-tuning of the head layers.

In contrast, Large models typically require fewer epochs but are trained with slightly lower learning rates. The batch size remains consistent across both model sizes, set at 32 for all tasks. Additionally, maximum sequence lengths are adjusted according to the task, with longer sequences allocated for more complex tasks like CoLA and QNLI.

\begin{table}
\centering
\resizebox{0.4\textwidth}{!}{%
\begin{tabular}{l|cc}
\toprule
\textbf{Hyperparameter} & \textbf{Medium} & \textbf{Large} \\
\midrule
Optimizer & \multicolumn{2}{c}{AdamW} \\
Learning Rate (SSH) & 2E-2 & 5E-2 \\
Learning Rate (FourierFT) & 2E-2 & 5E-2 \\
Learning Rate (VeRA) & 2E-2 & 5E-2 \\
Learning Rate (Head) & 2E-4 & 1E-4 \\
Batch Size & \multicolumn{2}{c}{128} \\
Weight Decay & 0.01 & 0.03 \\
$n_{SSH}$ &\multicolumn{2}{c}{750} \\
$n_{FourierFT}$ & \multicolumn{2}{c}{1000} \\
Scaling value $\alpha$ & \multicolumn{2}{c}{300} \\
Epochs & \multicolumn{2}{c}{5}  \\
Label Smooth & \multicolumn{2}{c}{0.1}  \\
LR Schedule & \multicolumn{2}{c}{Linear} \\
\bottomrule
\end{tabular}%
}
\caption{\small Hyperparameter settings on E2E benchmark}
\label{tab:nlgh}
\end{table}

\paragraph{Hyperparameter settings on E2E benchmark.} Tab.~\ref{tab:nlgh} details the hyperparameter configurations used for the medium and large models on the E2E benchmark. Both models are optimized using AdamW with a linear learning rate schedule. For the medium model, the learning rates for SSH and FourierFT are set to \(2E-2\), while for the large model, the learning rates are set to \(5E-2\). The head layers are fine-tuned with lower learning rates of \(2E-4\) for the medium model and \(1E-4\) for the large model. Both models employ a batch size of 128, with weight decay values of 0.01 for the medium model and 0.03 for the large model. The number of selected frequencies, \(n\), is set to 750 for SSH and 1000 for FourierFT, with the scaling factor \(\alpha\) fixed at 300 for both models. Additionally, label smoothing with a value of 0.1 is applied, and the models are trained for 5 epochs.

\begin{table}
\centering
\resizebox{0.45\textwidth}{!}{%
\begin{tabular}{l|cccc}
\toprule
\textbf{Hyperparameter} & \textbf{LoRA} & \textbf{FourierFT} & \textbf{SSH} & \textbf{VeRA}\\
\midrule
Optimizer & \multicolumn{4}{c}{AdamW} \\
Warmup Ratio & \multicolumn{4}{c}{0.06} \\
Batch Size & \multicolumn{4}{c}{4} \\
Acc. Steps & \multicolumn{4}{c}{4} \\
Epochs & \multicolumn{4}{c}{1} \\
$n$ &  -- & 1000 & 750 & --\\
Scaling Value $\alpha$ & 300.0 & 16.0 & 16.0 & 300.0\\
LR Schedule & \multicolumn{4}{c}{Linear} \\
Learning Rate & 3E-2 & 3E-3 & 3E-3 &  3E-3 \\
\bottomrule
\end{tabular}%
}
\caption{\small Hyperparameter settings for instruction-tuning configurations.}
\label{tab:hyperparamsIn}
\end{table}

\paragraph{Hyperparameter Setup for Instruction-Tuning.} Table~\ref{tab:hyperparamsIn} provides a summary of the key hyperparameters used for fine-tuning the LoRA, FourierFT, and SSH models. For all methods, the optimizer is AdamW, with a warmup ratio of 0.06. A batch size of 4 is used, along with gradient accumulation steps of 4 to ensure stability during training. The default training duration is 1 epoch, although in specific cases—such as the motivation example in the introduction and the ablation study in the supplementary material—2 epochs are used.

For SSH, the parameter \( n \) is set to 750. The scaling factor \( \alpha \) varies across methods: it is set to 300.0 for LoRA, 16.0 for FourierFT, and 16.0 for SSH. Learning rates are adjusted individually, with LoRA using 3E-2, while both FourierFT and SSH use a lower rate of 3E-3. All methods utilize a linear learning rate schedule.

\paragraph{Hyperparameter setup for image classification.}

\begin{table}
\centering
\resizebox{0.47\textwidth}{!}{%
\begin{tabular}{l|cccc}
\toprule
\textbf{Hyperparameter} & \textbf{CIFAR100} & \textbf{DTD} & \textbf{EuroSAT} &  \textbf{OxfordPets} \\
\midrule
Epochs & \multicolumn{4}{c}{10} \\
Optimizer & \multicolumn{4}{c}{AdamW} \\
LR Schedule & \multicolumn{4}{c}{Linear} \\
$n_{SSH}$ & \multicolumn{4}{c}{2250} \\
$n_{FourierFT}$ & \multicolumn{4}{c}{3000} \\
$\alpha$ & \multicolumn{4}{c}{300.0} \\
Learning Rate (SSH) & 2E-1 & 3E-1 & 2E-1 & 3E-1  \\
Learning Rate (FourierFT) & 2E-1 & 3E-1 & 2E-1 & 3E-1  \\
Learning Rate (VeRA) & 2E-1 & 3E-1 & 2E-1 & 3E-1  \\
Learning Rate (Head) & 7E-4 & 1E-3 & 8E-4 & 1E-3  \\
Weight Decay & 1E-4 & 7E-5 & 3E-4 & 8E-4 \\
\bottomrule
\end{tabular}%
}
\caption{\small Hyperparameter setup for image classification.}
\label{tab:SSH_image}
\end{table}

Tab.~\ref{tab:SSH_image} outlines the hyperparameter configurations used for fine-tuning on the CIFAR100, DTD, EuroSAT, and OxfordPets datasets for image classification tasks. The table provides the common settings applied across all datasets, including the use of the AdamW optimizer, a linear learning rate schedule, and a consistent training duration of 10 epochs. The number of frequency components ($n$) is set to 2250 for SSH and 3000 for FourierFT across all datasets.

For both SSH and FourierFT, the learning rate varies slightly depending on the dataset, ranging from 2E-1 to 3E-1, while the learning rate for the classification head lies between 7E-4 and 1E-3. The weight decay is also tuned per dataset, with values between 7E-5 and 1E-4 for DTD and CIFAR100, and slightly higher at 3E-4 and 8E-4 for EuroSAT and OxfordPets, respectively.

\end{document}